\RequirePackage{fix-cm}
\documentclass[smallcondensed]{svjour3}     
\smartqed  
\usepackage{graphicx}
\journalname{Neural Processing Letters}
\usepackage{csquotes}
\usepackage{amsmath}
\usepackage{hyperref}

\smartqed  
\makeatletter
\def\cl@chapter{\@elt {theorem}}
\makeatother

\usepackage{cleveref} 
\usepackage{enumitem}
\usepackage[ruled,linesnumbered]{algorithm2e}
\usepackage{setspace}
\usepackage{siunitx}
\usepackage{diagbox}
\usepackage{multirow}
\usepackage{hhline}
\usepackage{amssymb}
\usepackage[center]{subfigure}
\usepackage{academicons}
\usepackage{xcolor}

\usepackage{amsmath,amsfonts,bm}

\def\eqref#1{equation~\ref{#1}}

\def\1{\bm{1}}

\def\eps{{\epsilon}}

\def\vmu{{\bm{\mu}}}
\def\vtheta{{\bm{\theta}}}

\def\vg{{\bm{g}}}

\def\vv{{\bm{v}}}

\def\vx{{\bm{x}}}

\def\vz{{\bm{z}}}

\def\mD{{\bm{D}}}

\def\mP{{\bm{P}}}

\def\mX{{\bm{X}}}

\def\mSigma{{\bm{\Sigma}}}

\DeclareMathAlphabet{\mathsfit}{\encodingdefault}{\sfdefault}{m}{sl}
\SetMathAlphabet{\mathsfit}{bold}{\encodingdefault}{\sfdefault}{bx}{n}

\newcommand{\E}{\mathbb{E}}

\newcommand{\Cov}{\mathrm{Cov}}

\DeclareMathOperator*{\argmax}{arg\,max}

\DeclareMathOperator{\Tr}{Tr}

\newcommand{\wrt}{w.r.t.\ }
\newcommand{\orcid}[1]{\href{https://orcid.org/#1}{\includegraphics[height=1em]{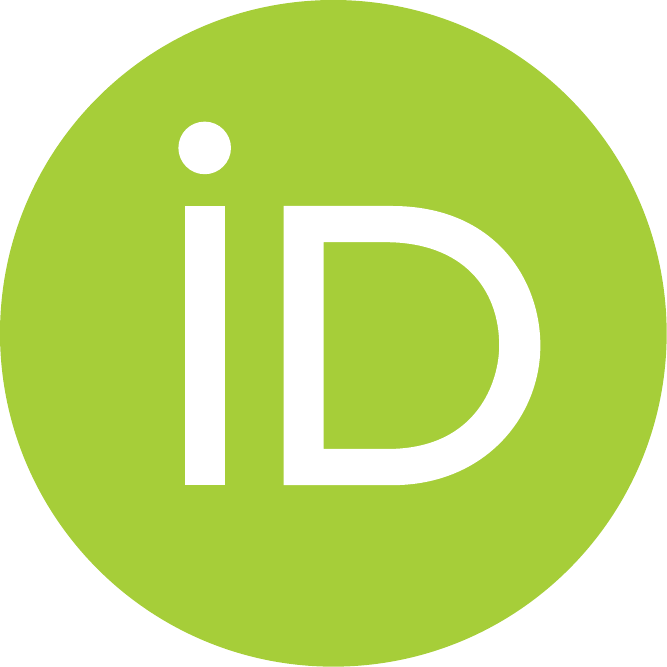}}}

\crefname{algorithm}{Alg.}{Algs.}
\begin{document}

\title{Gradient-based training of Gaussian Mixture Models for High-Dimensional Streaming Data}
\titlerunning{Gradient-based training of GMMs for High-Dimensional Streaming Data}        
\author{Alexander Gepperth\orcid{0000-0003-2216-7808} \and Benedikt Pf{\"u}lb\orcid{0000-0002-0108-1936}}
\authorrunning{Alexander Gepperth and Benedikt Pf{\"u}lb}
\institute{Fulda University of Applied Sciences \at Leipziger Str.\ 123, 36037 Fulda \\ \email{\{alexander.gepperth,benedikt.pfuelb\}@cs.hs-fulda.de}}

\date{Accepted: May 24, 2021} 

\maketitle
\begin{abstract}
We present an approach for efficiently training Gaussian Mixture Model (GMM) by Stochastic Gradient Descent (SGD) with non-stationary, high-di\-me\-nsion\-al streaming data.
Our training scheme does not require data-driven parameter initialization (e.g., k-means) and can thus be trained based on a random initialization.
Furthermore, the approach allows mini-batch sizes as low as $1$, which are typical for streaming-data settings.
Major problems in such settings are undesirable local optima during early training phases and numerical instabilities due to high data dimensionalities.
We introduce an adaptive annealing procedure to address the first problem, whereas numerical instabilities are eliminated by using an exponential-free approximation to the standard GMM log-likelihood.
Experiments on a variety of visual and non-visual benchmarks show that our SGD approach can be trained completely without, for instance, k-means based centroid initialization.
It also compares favorably to an online variant of Expectation-Maximization (EM) -- stochastic EM (sEM), which it outperforms by a large margin for very high-dimensional data.
\end{abstract}

\section{Introduction}
This contribution focuses on Gaussian Mixture Model (GMM), which represent a probabilistic unsupervised model for clustering and density estimation, allowing sampling and outlier detection.
GMMs have been used in a wide range of scenarios, see \cite{Melnykov2010}.
Commonly, free parameters of a GMM are estimated using the Expectation-Maximization (EM) algorithm~\cite{Dempster1977}, which does not require learning rates and automatically enforces all GMM constraints.
A popular online variant is stochachstic EM \cite{Cappe2009}, which can be trained mini-batch wise and is thus more suited for large datasets or streaming data.

\subsection{Motivation}
Intrinsically, EM is a batch-type algorithm.
Therefore, memory requirements can become excessive for large datasets.
In addition, streaming-data scenarios require data samples to be processed one by one, which is impossible for a batch-type algorithm.
Moreover, data statistics may be subject to changes over time (concept drift/shift), to which the GMM should adapt.
In such scenarios, an online, mini-batch type of optimization such as SGD is attractive since it can process samples one by one, has modest, fixed memory requirements, and can adapt to changing data statistics.

\subsection{Related Work}
\textbf{Online EM} is a technique for performing EM mini-batch wise, allowing to process large datasets.
One branch of previous research~\cite{Newton1986,Lange1995,Chen2018} has been devoted to the development of stochastic Expectation-Maximization (sEM) algorithms that reduce to the original EM method in the limit of large batch sizes.
The  variant presented in \cite{Cappe2009} is widely used due to its simplicity and efficiency for large datasets.
Such approaches come at the price of additional hyper-parameters (e.g., step size, mini-batch size, step size reduction), thus, removing a key advantage of EM over SGD.
Another approach is to modify the EM algorithm itself by, e.g., including heuristics for adding, splitting and merging centroids \cite{Vlassis2002,Engel2010,Pinto2015,Cederborg2010,Song2005,Kristan2008,Vijayakumar2005}.
This allows GMM-like models to be trained by presenting one sample after another.
The models work well in several application scenarios, but their learning dynamics are impossible to analyze mathematically.
They also introduce a high number of parameters.
Apart from these works, some authors avoid the issue of extensive datasets by determining smaller \enquote{core sets} of representative samples and performing vanilla EM~\cite{Feldman2011}.
\\
\textbf{SGD for training GMM} has, as far as we know, been recently treated only in \cite{Hosseini2015,hosseini2019}.
In this body of work, GMM constraint enforcement is ensured by using manifold optimization techniques and re-parameterization/regularization, thereby introducing additional hyper-parameters.
The issue of local optima is side-stepped by a k-means type centroid initialization, and the used datasets are low-dimensional ($36$ dimensions).
\\
\textbf{Annealing and Approximation approaches for GMMs} were proposed in \cite{Verbeek2005,Pinheiro1995,Ormoneit1998,Dognin2009}.
However, the regularizers proposed in \cite{Verbeek2005,Ormoneit1998} significantly differ from our scheme.
GMM log-likelihood approximations, similar to the one used here, are discussed in, e.g., \cite{Pinheiro1995} and \cite{Dognin2009}, but only in combination with EM training.
A similar \enquote{hard assignment} approximation is performed in \cite{van2014factoring}.
\\
\textbf{GMM Training in High-Dimensional Spaces} is discussed in several publications: A conceptually very interesting procedure is proposed in \cite{Ge2015}.
It exploits the properties of high-dimensional spaces in order to achieve learning with a number of samples that is polynomial in the number of Gaussian components.
This is difficult to apply in streaming settings, since higher-order moments need to be estimated beforehand, and also because the number of samples is usually unknown.
Training GMM-like lower-dimensional factor analysis models by SGD on high-dimensional image data is successfully demonstrated in \cite{Richardson2018}.
They avoid numerical issues, but, again, sidestep the local optima issue by using k-means initialization.
The numerical issues associated with log-likelihood computation in high-dimensional spaces are generally mitigated by using the \enquote{logsumexp} trick~\cite{Nielsen2016}, which is, however, insufficient for ensuring numerical stability for particularly high-dimensional data, such as images.

\subsection{Goals and Contributions}
The goals of this article are to establish GMM training by SGD as a simple and scalable alternative to sEM in streaming scenarios with potentially high-dimensional data.
The main novel contributions are:
\vspace{-0.5em}
\begin{itemize}
	\item a proposal for numerically stable GMM training by SGD that outperforms sEM for high data dimensionalities,
	\item an automatic annealing procedure that ensures SGD convergence without prior knowledge of the data (\textbf{no} k-means initialization) which is beneficial for streaming data,
	\item a computationally efficient method for enforcing all GMM constraints in SGD,
	\item a convergence proof for the annealing procedure.
\end{itemize}
Additionally, we provide a TensorFlow implementation.\footnote{\href{https://gitlab.cs.hs-fulda.de/ML-Projects/sgd-gmm}{https://gitlab.cs.hs-fulda.de/ML-Projects/sgd-gmm}}
\section{Gaussian Mixture Models}\label{sec:sgd}
GMMs are probabilistic models that intend to explain the observed data $X$\,$=$\,$\{\vx_n\}$ by expressing their density as a weighted mixture of $K$ Gaussian component densities $\mathcal{N}(\vx;\vmu_k,\mP_k)$\,$\equiv$\,$\mathcal{N}_k(\vx)$: $p(\vx)$\,$=$\,$\sum_k^K \pi_k \mathcal{N}_k(\vx)$.
We parameterize Gaussian densities by precision matrices $\mP_k=\mSigma_k^{-1}$ instead of covariances $\mSigma_k$.
GMM training optimizes the (incomplete) log-likelihood
\begin{equation}
	\mathcal{L} = \E_n\left[ \log \sum_k \pi_k \mathcal{N}_k(\vx_n)\right].
	\label{eqn:loglik}
\end{equation}
\subsection{GMM Constraint Enforcement for SGD}\label{sec:sgd-constraints}
GMMs require the mixture weights to be normalized: $\sum_k \pi_k$\,$=$\,$1$ and the precision matrices to be positive definite: $\vx^\top \mP_k \vx$\,$\ge$\,$0$\ $\forall \vx$.
These constraints must be explicitly enforced after each SGD step:
\\
\textbf{Weights} $\pi_k$ are adapted according to \cite{Hosseini2015}, which replaces them by other free parameters $\xi_k$ from which the $\pi_k$ are computed so that normalization is ensured:
\begin{equation}
	\pi_k = \frac{\exp(\xi_k)}{\sum_j \exp(\xi_j)}.
	\label{eqn:pi_norm}
\end{equation} 
\\
\textbf{Diagonal precision matrices} are re-parameterized as $\mP_k$\,$=$\,$\mD_k^2$, with diagonal matrices $\mD_k$ (Cholesky decomposition).
They are, therefore, guaranteed to be positive definite.
Hence, $\det$\,$\mSigma_k$\,$=$\,$\det$\,$\mP^{-1}_k$\,$=$\,$\left(\det(\mD^2_k)\right)^{-1}$\,$=$\,$\big(\Tr(\mD_k)\big)^{-2}$ can be computed efficiently.
Since we are dealing with high-dimensional data, precision matrices are always taken to be diagonal, since full matrices would be prohibitive \wrt memory consumption and the number of free parameters.
\par\noindent\textbf{Full precision matrices} are treated here for completeness' sake, since they are infeasible for high-dimensional data.
We represent them as a spectral decomposition into eigenvectors $\vv_i$ and eigenvalues $\lambda_i^2$: $\mP_k = \sum_i \lambda_i^2 \vv_i \vv_i^\top$, which ensures positive-definiteness.
This can be seen from $\det$\,$\mSigma_k$\,$=$\,$\det$\,$\mP^{-1}_k$\,$=$\,$\prod_i \lambda_i^{-2}$.
In order to maintain a correct representation of eigenvectors, these have to be orthonormalized after each SGD step.
\subsection{Max-Component Approximation for GMM}\label{sec:max-comp}
The log-likelihood \cref{eqn:loglik} is difficult to optimize by SGD due to numerical problems (mainly underflows and resulting divisions by zero) for high data dimensionalities.
This is why we intend to find a lower bound that we can optimize instead.
A simple scheme is given by
\begin{equation}
	\begin{split}
		\mathcal{L}\! & =\!\E_n\!\left[ \log \sum_k \pi_k \mathcal{N}_k(\vx_n)\right]\!\ge\!\E_n\Big[ \log \text{max}_k \big(\pi_k \mathcal{N}_k(\vx_n) \big) \Big] \\
		              & =\!\hat{\mathcal{L}}\!=\!\E_n\Big[ \log \big(\pi_{k^*} \mathcal{N}_{k^*}(\vx_n) \big) \Big]
	\end{split}
	\label{eqn:max}
	\raisetag{20pt}
\end{equation}
where ${k^*}$\,$=$\,$\argmax_k \pi_k\mathcal{N}_k(\vx_n)$.
This is what we call the \textit{max-component approximation} of \cref{eqn:max}. 
In contrast to the lower bound that is constructed for EM-type algorithms, our bound is usually not tight.
Nevertheless, we will demonstrate later that it is a very good approximation when data are high-dimensional.
The advantage of $\hat{\mathcal{L}}$ is the elimination of exponentials causing numerical instabilities.
The \enquote{logsumexp} trick is normally employed with GMMs to rectify this by factoring out the largest component probability $\mathcal{N}_{k^*}$.
This mitigates but does not avoid numerical problems when distances are high, a common occurrence for high data dimensions.
To give an example: we normalize the component probability $\mathcal{N}_k$\,$=$\,$e^{-101}$ (using 32-bit floats) by the highest probability $\mathcal{N}_{k^*}$\,$=$\,$e^{3}$, and we obtain $\frac{\mathcal{N}_k}{\mathcal{N}_{k^*}}$\,$=$\,$e^{-104}$, which produces an underflow.
\subsection{Undesirable Local Optima in SGD Training}\label{sec:local}
\begin{figure}[htb!]
	\centering
	\includegraphics[width=0.4\textwidth]{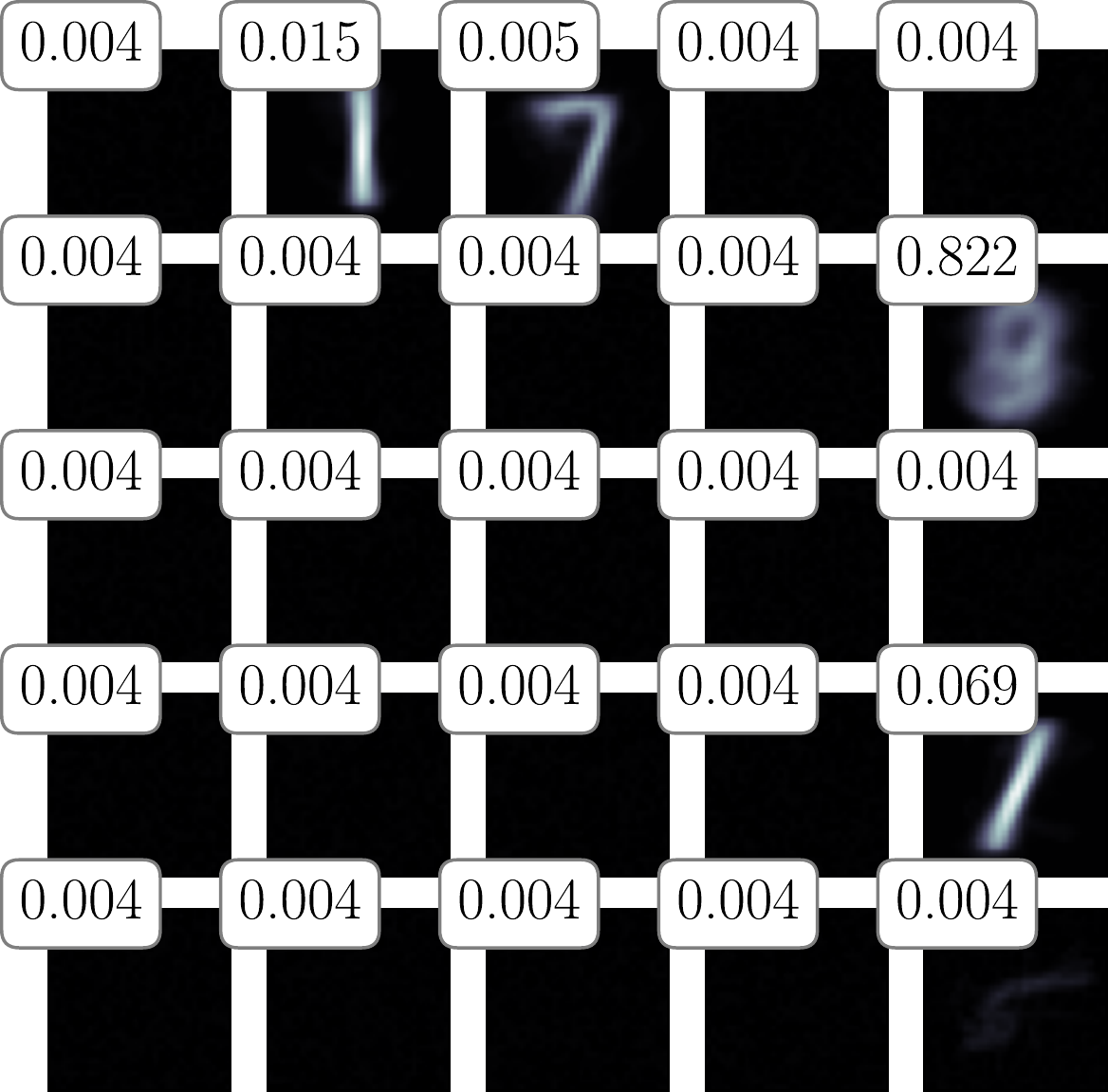}
	\caption{A sparse-component-solution with superimposed component weights $\pi_k$, obtained when performing naive SGD on MNIST.}
	\label{fig:sparse_solution}
\end{figure}

A crucial issue when optimizing $\hat{\mathcal{L}}$ (and indeed $\mathcal L$ as well) by SGD without k-means initialization concerns undesirable local optima.
Most notable are the \textbf{sin\-gle/sparse-component solutions}, see \cref{fig:sparse_solution}.
They are characterized by one or several components $\{k_i\}$ having large weights, with centroid and precision matrices given by the mean and covariance of a significant subset $\mX_{k_i}$\,$\subset$\,$\mX$ of the data $\mX$: $\pi_{{k_i}}$\,$\gg$\,$0$, $\vmu_{{k_i}}$\,$=$\,$\E[\mX_{k_i}]$, $\mSigma_{{k_i}}$\,$=$\,$\Cov(\mX_{k_i})$, whereas the remaining components $k$ are characterized by $\pi_{{k}}$\,$\approx$\,$0$, $\vmu_{{k}}$\,$=$\,$\vmu(t$\,$=$\,$0)$, $\mP_{{k}}$\,$=$\,$\mP(t\!=\!0)$.
Thus, these unconverged components are almost never Best Matching Unit (BMU) $k^*$.
The max-operation in $\hat{\mathcal{L}}$ causes gradients like $\frac{\partial \hat{\mathcal{L}}}{\partial \vmu_{k}}$ to contain $\delta_{kk^*}$:
\begin{equation}
	\begin{split}
		\frac{\partial \hat{\mathcal{L}}}{\partial \vmu_{k}} & = \E_n\left[  \mP_k\left(\vx_{n}-\vmu_{k}\right)\delta_{kk^*}\right]                             \\[.5em]
		\frac{\partial \hat{\mathcal{L}}}{\partial \mP_{k}}  & = \E_n\left[ \left( (\mP_k)^{-1} - (\vx_n - \vmu_k)(\vx_n - \vmu_k)^\top\right)\delta_{kk^*}\right] \\[.5em]
		\frac{\partial \hat{\mathcal{L}}}{\partial \pi_{k}}  & = \pi_k^{-1}\E_n\left[\delta_{kk^*}\right].
	\end{split}
	\label{eqn:mcgrads}
	\raisetag{20pt}
\end{equation}
This implies that the gradients are non-zero only for the BMU $k^*$.
Thus, the gradients of unconverged components vanish, implying that they remain in their unconverged state.

\subsection{Annealing Procedure for Avoiding Local Optima}\label{sec:sgd-reg}
\begin{figure}[htb!]
	\centering
	\includegraphics[width=.9\linewidth]{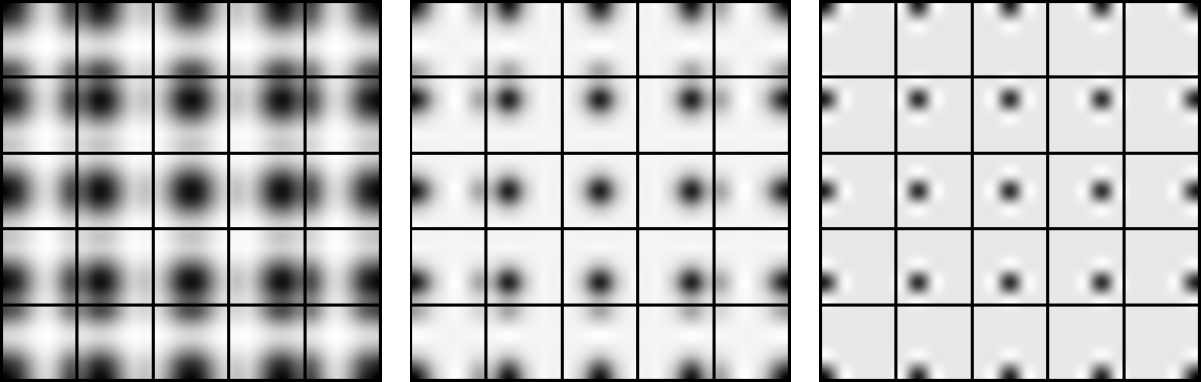}\label{fig:iter2}
	\caption{
		Visualization of Gaussian smoothing filters $\vg_k$, of width $\sigma$, used in annealing for three different values of $\sigma$.
		The $\vg_k$ are placed on a 2D grid, darker pixels indicate larger values.
		Over time, $\sigma(t)$ is reduced (middle and right pictures) and the Gaussians approach a delta peak, thus, recovering the original, non-annealed loss function.
		Note that the grid is considered periodic in order to avoid boundary effects, so the $\vg_k$ are themselves periodic.
	}
	\label{fig:conv_mask}
\end{figure}

Our approach for avoiding sparse-component solutions is to punish their characteristic response patterns by replacing $\hat{\mathcal{L}}$ by the \textit{smoothed max-component log-likelihood} $\hat{\mathcal{L}}^\sigma$:
\begin{equation}
	\begin{split}
		\hat{\mathcal{L}}^\sigma & = \E_n \text{max}_k \Bigg[ \sum_j \vg_{kj}(\sigma) \log \Big(\pi_j \mathcal{N}_j(\vx_n)\Big)\Bigg] \\
		                         & = \E_n \sum_j \vg_{k^*j}(\sigma) \log \Big(\pi_j \mathcal{N}_j(\vx_n)\Big).
	\end{split}
	\label{eqn:conv}
\end{equation}
Regarding its interpretation, we are assuming that the $K$ GMM components are arranged in a quadratic 2D grid of size $\sqrt{K}$\,$\times$\,$\sqrt{K}$.
Equally, each $\vg_k$ is interpreted as 2D grid of size $\sqrt{K}$\,$\times$\,$\sqrt{K}$, (see \cref{fig:conv_mask}), with
values given by a periodically continued 2D Gaussian centered on component $k$.
With this interpretation, 
\Cref{eqn:conv} represents a 2D convolution with periodic boundary conditions (in the sense used in image processing) of the $\log\left(\pi_k\mathcal{N}_k(\vx)\right)$ by a smoothing filter whose width is controlled by $\sigma$.
Thus, \cref{eqn:conv} is maximized if the log-probabilities follow a uni-modal Gaussian profile of spatial variance $\sim$\,$\sigma^2$, which heavily punishes single/sparse-component solutions that have a locally delta-like response. 
A 1D grid for annealing, together with 1D smoothing filters, was verified to fulfill this purpose as well. 
We chose 2D because it allows for an easier visualization while incurring an identical computational cost.
\smallskip\\
Annealing starts with a large value of $\sigma(t)$\,$=$\,$\sigma_0$ and reduces it over time to an asymptotic small value of $\sigma$\,$=$\,$\sigma_{\infty}$, thus, smoothly transitioning from $\hat{\mathcal{L}}^\sigma$ in \cref{eqn:conv} into $\hat{\mathcal{L}}$ in \cref{eqn:max}.
\smallskip\\
\textbf{Annealing Control} regulates the decrease of $\sigma$.
This quantity defines an effective upper bound on $\hat{\mathcal{L}}^\sigma$ (see \cref{ssec:proof_annealing_upper_bound} for a proof).
An implication is that the loss will be stationary once this bound is reached, which we consider a suitable indicator for reducing $\sigma$.
We implement an annealing control that sets $\sigma$\,$\leftarrow$\,$0.9\sigma$ whenever the loss is considered sufficiently stationary.
Stationarity is detected by maintaining an exponentially smoothed average $\ell(t)$\,=\,$(1-\alpha)\ell(t-1)+\alpha\hat{\mathcal{L}}^\sigma(t)$ on time scale $\alpha$.
Every $\frac{1}{\alpha}$ iterations, we compute the fractional increase of $\hat{\mathcal{L}}^\sigma$ as
\begin{equation}
	\Delta = \frac{\ell(t) - {\ell}(t-\alpha^{-1})}{{\ell}(t-\alpha^{-1}) -\hat{\mathcal{L}}^\sigma(t=0)}
	\label{eqn:delta}
\end{equation}
and consider the loss stationary iff $\Delta$\,$<$\,$\delta$ (the latter being a free parameter).
The choice of the time constant for smoothing  $\hat{\mathcal{L}}^\sigma$ is outlined in the following section.
\subsection{Training Procedure for SGD}\label{sec:sgd-full}\label{sec:train-sgd}
Training GMMs by SGD is performed by maximizing the smoothed max-com\-po\-nent log-likelihood $\hat{\mathcal{L}}^{\sigma}$ from \cref{eqn:conv}.
At the same time, we enforce the constraints on the component weights and covariances as described in \cref{sec:sgd-constraints} and transition from $\hat{\mathcal{L}}^\sigma$ into $\hat{\mathcal{L}}$ by annealing (see \cref{sec:sgd-reg}). 
SGD requires a learning rate $\epsilon$ to be set, which in turn determines the parameter $\alpha$ (see \cref{sec:sgd-reg}) as $\alpha$\,$=$\,$\epsilon$ since stationarity detection should operate on a time scale similar to that of SGD. 
The diagonal matrices $\mD_k$ are initialized to $D_\text{max}I$ and are clipped after each iteration so that diagonal entries remain in the range $[0,D_\text{max}^2]$.
This is necessary to avoid excessive growth of precisions for data entries with vanishing variance, e.g., pixels that are always black.
Weights are uniformly initialized to $\pi^i$\,$=$\,$\frac{1}{K}$, centroids in the range $[-\mu^i,+\mu^i]$ (see \autoref{algo:sgd_gmm_training} for a summary).
Please note that our SGD approach requires no centroid initialization by \mbox{k-means}, as it is recommended when training GMMs with (s)EM.
We discuss and summarize good practices for choosing hyper-parameters in \cref{sec:discussion}.
\makeatletter 
\patchcmd{\@algocf@start}
{-1.5em}
{0pt}
{}{}
\makeatother
\vspace{-1em}
\IncMargin{1.5em}
\begin{algorithm}[h!]
	\small
	\SetInd{1.em}{.1em}
	\SetEndCharOfAlgoLine{}
	\SetAlgoVlined 
	\SetAlgorithmName{\normalfont{Algorithm}}{}{}
	\SetAlgoCaptionSeparator{\normalfont{:}}
	\Indm
	\KwData{initializer values: $\mu^i$, $K$, $\eps_0$/$\eps_\infty$, $\sigma_0$/$\sigma_\infty$, $\delta$ and data $\mX$} 
	\KwResult{trained GMM model}
	\Indp\Indpp
	$\vmu$    $\leftarrow$ $\mathcal{U}(-\mu^i, +\mu^i)$,\,
	$\pi$     $\leftarrow$ $1 / K$,\,                     
	$\mP$     $\leftarrow$ $I D_\text{max}$,\,                                    
	$\sigma$  $\leftarrow$ $\sigma_0$,\,
	$\eps$    $\leftarrow$ $\eps_0$                                                                                                                              \\ 
	\ForAll(\hfill // training loop){t $<$ T}{                                             
		$\vg(t)$ $\leftarrow$ create\_annealing\_mask($\sigma$,$t$)                                             \hfill // see \cref{sec:sgd-reg}                 \\
		$\vmu(t)$ $\leftarrow$ $\epsilon\frac{\partial\hat{\mathcal{L}}^\sigma}{\partial\vmu}\!+\!\vmu(t$-$1)$, \hfill // SGD updates                            \\                                                   
		$\mP(t)$  $\leftarrow$	$\epsilon\frac{\partial\hat{\mathcal{L}}^\sigma}{\partial\mP}\!+\!\mP(t$-$1)$,                                                   \\ 
		$\pi(t)$  $\leftarrow$ $\epsilon\frac{\partial\hat{\mathcal{L}}^\sigma}{\partial\pi}\!+\!\pi(t$-$1)$                                                     \\                                                   
		$\mP(t)$  $\leftarrow$ precisions\_clipping($\mP$, $D_\text{max}$)                                     \hfill //see \cref{sec:train-sgd}                 \\
		$\pi(t)$  $\leftarrow$ normalization($\pi(t)$)                                                         \hfill //see \cref{eqn:pi_norm}                   \\
		$\ell(t)$ $\leftarrow$ $(1\!-\!\alpha) \ell(t\!-\!1)\!+\!\alpha \hat{\mathcal{L}}^\sigma(\vx(t))$      \hfill // sliding likelihood                      \\
		\If(\hfill // see \cref{sec:sgd-reg}){annealing update iteration}{
			\lIf(){$\Delta < \delta$}{                                                                         \hfill // $\Delta$ see \cref{eqn:delta}           \\
				\hspace*{1em}$\sigma(t)$ $\leftarrow$ $0.9\sigma(t\!-\!1)$,\,$\eps(t)$ $\leftarrow$ $0.9\eps(t\!-\!1)$
			}
		}
	}
	\caption{Steps of SGD-GMM training.}
	\label{algo:sgd_gmm_training}
\end{algorithm}
\DecMargin{1.5em}
\subsection{Proof that Annealing is Convergent}\label{ssec:proof_annealing_upper_bound}
We assume that, for a fixed value of $\sigma$, SGD optimization has reached a stationary point where the derivative \wrt all GMM parameters is $0$ on average.
In this situation, we claim that decreasing $\sigma$ will always increase the loss.
If true, this would show that $\sigma$ defines an effective upper bound for the loss.
For this to be consistent, we have to show that the loss gradient \wrt $\sigma$ vanishes as $\sigma\!\rightarrow\!0$: as the annealed loss approaches the original one, decreases of $\sigma$ have less and less effects.
\par\noindent \textbf{Proposition} 
The gradient $\frac{\partial \hat {\mathcal {L}}^{\sigma}}{\partial \sigma}$ is strictly positive for $\sigma\!>\!0$ 
\par\noindent \textbf{Proof}
For each sample, the 2D profile of $\log(\pi_k \mathcal{N}_k)$\,$\equiv$\,$f_k$ is assumed to be centered on the best-matching component $k^*$ and depends on the distance from it as a function of $||k-k^*||$.
We thus have $f_k$\,$=$\,$f_k(r)$ with $r$\,$\equiv$\,$||k-k^*||$.
Passing to the continuous domain, the indices in the Gaussian \enquote{smoothing filter} $g_{k^*k}$ become continuous variables, and we have $g_{k^*k}\rightarrow g(||k-k^*||,\sigma)\!\equiv\!g(r,\sigma)$. 
Similarly, $f_k(r)\!\rightarrow\!f(r)$.
Using 2D polar coordinates, the smoothed max-component likelihood $\hat{\mathcal{L}}^\sigma$ becomes a polar integral around the position of the best-matching component: $\hat{\mathcal{L}}^\sigma$\,$\sim$\,$\int_{\mathbb{R}^2} g(r,\sigma)f(r)drd\phi$.
It is trivial to show that for the special case of a constant log-probability profile, i.e., $f(r)$\,$=$\,$L$, $\mathcal{L}^\sigma$, does not depend on $\sigma$ because Gaussians are normalized, and that the derivative \wrt $\sigma$ vanishes:
\begin{equation}
	\begin{split}
		\frac{d\hat{\mathcal{L}}^\sigma}{d\sigma} & \!\sim\!\int_0^\infty\!dr \Big(\frac{r^2}{\sigma^2}\!-\!1\Big) \exp\Big(\!-\frac{r^2}{2\sigma^2}\Big)L                                                                                \\ 
		& \!=\!L\int_0^\sigma\!dr \Big(\frac{r^2}{\sigma^2}\!-\!1\Big) \exp\Big(\!-\frac{r^2}{2\sigma^2}\Big)\!-\!L\!\int_\sigma^\infty\!\Big(\frac{r^2}{\sigma^2}\!-\!1\Big) \exp\Big(\!-\frac{r^2}{2\sigma^2}\Big) \\
		& \!\equiv\!L\mathcal{N}\!-\!L\mathcal{P}
	\end{split}
	\raisetag{53pt}
\end{equation}
where we have split the integral into parts where the derivative \wrt $\sigma$ is negative ($\mathcal{N}$) and positive ($\mathcal{P}$).
We know that $\mathcal{N} = \mathcal{P}$ since the derivative must be zero for a constant function $f(r)=L$ due to the fact that Gaussians are normalized to the same value regardless of $\sigma$.
\par
For a function $f(r)$ that satisfies $f(r)\!>\!L \forall r\in [0,\sigma[ $ and $f(r)\!<\!L\forall r\in]\sigma, \infty[$, the inner and outer parts of the integral behave as follows:
\begin{equation}
	\begin{split}
		\tilde{\mathcal{N}} &\!=\!\int_0^\sigma\!g(r)\Big(\frac{r^2}{\sigma^2}\!-\!1\Big) f(r)\!<\!\int_0^\sigma\!g(r)\Big(\frac{r^2}{\sigma^2}\!-\!1\Big) L\!=\!L\mathcal{N}          \\
		\tilde{\mathcal{P}} &\!=\!\int_\sigma^\infty\!g(r)\Big(\frac{r^2}{\sigma^2}\!-\!1\Big) f(r)\!<\!\int_\sigma^\infty\!g(r)\Big(\frac{r^2}{\sigma^2}\!-\!1\Big) L\!=\!L\mathcal{P}
	\end{split}
	\raisetag{32pt}
\end{equation}
since $f(r)$ is minorized/majorized by $L$ by assumption, and the contributions in both integrals have the same sign for the whole domain of integration.
Thus, it is shown that
\begin{equation}
	\frac{d\hat{\mathcal{L}}^\sigma}{d\sigma} = \tilde{\mathcal{N}} - \tilde{\mathcal{P}} < L{\mathcal{N}} - L{\mathcal{P}} = 0
\end{equation}
for $\sigma$\,$>$\,$0$ and, furthermore, that this derivative is zero for $\sigma$\,$=$\,$0$ because $\hat{\mathcal{L}}^\sigma$ no longer depends on $\sigma$ for this case.
\smallskip\\
Taking everything into consideration, in a situation where the log-likelihood $\hat{\mathcal{L}}^\sigma$ has reached a stationary point for a given value of $\sigma$, we have shown that:
\begin{itemize}
	\item the value of $\hat{\mathcal{L}}^\sigma$ depends on $\sigma$,
	\item without changing the log-probabilities, we can increase $\hat{\mathcal{L}}^\sigma$ by reducing $\sigma$, assuming that the log-probabilities are mildly decreasing around the BMU,
	\item increasing $\hat{\mathcal{L}}^\sigma$ works as long as $\sigma$\,$>$\,$0$.
	At $\sigma\!=\!0$ the derivative becomes $0$.
\end{itemize}
Thus, $\sigma$ indeed defines an upper bound to $\hat{\mathcal{L}}^\sigma$ which can be increased by decreasing $\sigma$.
The assumption of log-probabilities that decrease, on average, around the BMU is reasonable, since such a profile maximizes $\hat{\mathcal{L}}^\sigma$.
All functions $f(r)$ that, e.g., decrease monotonically around the BMU, fulfill this criterion, whereas the form of the decrease is irrelevant.
\subsection{Training Procedure for sEM}\label{sec:train-sem}
We use sEM proposed by \cite{Cappe2009} as a reference point to which we compare our SGD approach.
We choose the step size of the form $\rho_{t}$\,$=$\,$\rho_0(t+1)^{-0.5+\alpha}$, with $\alpha$\,$\in$\,$[0,0.5]$, $\rho_0$\,$<$\,$1$ and enforce $\rho(t)$\,$\ge$\,$\rho_\infty$.
Values for these parameters are determined via a grid search in the ranges $\rho_0$\,$\in$\,$\{0.01,0.05,0.1\}$, $\alpha$\,$\in$\,$\{0.01,0.25,0.5\}$ and $\rho_\infty$\,$\in$\,$\{0.01,0.001,0.0001\}$.
Each sEM iteration uses a batch size $B$.
Initial accumulation of sufficient statics is conducted for $10\%$ of an epoch.
Parameter initialization and clipping of precisions is performed just as for SGD, see \cref{sec:sgd-full}.
\subsection{Comparing SGD and sEM}\label{sec:exp:fair}
Since sEM optimizes the log-likelihood $\mathcal{L}$, whereas SGD optimizes the annealed approximation $\hat{\mathcal{L}}^\sigma$, the comparison of these measures should be considered carefully.
We claim that the comparison is fair assuming that \textbf{i)} SGD annealing has converged and \textbf{ii)} GMM responsibilities are sharply peaked so that a single component has a responsibility of $\approx$\,$1$.
It follows from \textbf{i)} that $\hat{\mathcal{L}}^\sigma$\,$\approx$\,$\hat{\mathcal{L}}$ and \textbf{ii)} implies that $\hat{\mathcal{L}}$\,$\approx$\,$\mathcal{L}$.
Condition \textbf{ii)} is usually satisfied to high precision especially for high-dimensional inputs: if it is not, the comparison is biased in favor of sEM, since $\mathcal{L}$\,$>$\,$\hat{\mathcal{L}}$ by definition.
\section{Experiments}\label{sec:exp}
Unless stated otherwise, the experiments in this section will be conducted with the following parameter values for sEM and SGD (where applicable): mini-batch size $B$\,$=$\,$1$, $K$\,$=$\,$8$\,$\times$\,$8$, $\mu^i$\,$=$\,$0.1$, $\sigma_0$\,$=$\,$2$, $\sigma_\infty$\,$=$\,$0.01$, $\epsilon$\,$=$\,$0.001$, $D_{\text{max}}$\,$=$\,$20$.
Each experiment is repeated $10$ times with identical parameters but different random seeds for parameter initialization.
See \cref{sec:discussion} for a justification of these choices.
Due to input dimensionality, all precision matrices are assumed to be diagonal.
The training/test data comes from the datasets shown below (see \cref{sec:datasets}).

\subsection{Datasets}\label{sec:datasets}
We use a variety of different image-based datasets, as well as a non-image dataset for evaluation purposes.
All datasets are normalized to the $[0,1]$ range.
\\
\textbf{MNIST}~\cite{LeCun1998} contains gray-scale images, which depict handwritten digits from 0 to 9 in a resolution of $28$\,$\times28$ pixels -- the common benchmark for computer vision systems and classification problems.
\\
\textbf{SVHN}~\cite{wang2012end} contains color images of house numbers ($0$-$9$, resolution $32$\,$\times$\,$32$).
\\
\textbf{FashionMNIST}~\cite{Xiao2017} contains grayscale images of 10 clothing categories and is considered as a more challenging classification task compared to MNIST.
\\
\textbf{Fruits 360}~\cite{Muresan2017} consists of color pictures ($100$\,$\times$\,$100$\,$\times$\,$3$ pixels) showing different types of fruits.
The ten best-represented classes are selected.
\\
\textbf{Devanagari}~\cite{Acharya2015} includes grayscale images of handwritten Devanagari letters with a resolution of $32$\,$\times32$ pixels -- the first 10 classes are selected.
\\
\textbf{NotMNIST}~\cite{Yaroslav2011} is a grayscale image dataset (resolution $28$\,$\times$\,$28$ pixels) of letters from \textit{A} to \textit{J} extracted from different publicly available fonts.
\\
\textbf{ISOLET}~\cite{Cole1990} is a non-image dataset containing $7\,797$ samples of spoken letters recorded from $150$ subjects.
Each sample is encoded and is represented by $617$ float values.

\subsection{Robustness of SGD to Initial Conditions}\label{sec:exp:init}
Here, we train GMM for three epochs on classes 1 to 9 for each dataset.
We use different random and non-random initializations of the centroids and compare the final log-likelihood values.
Random centroid initializations are parameterized by $\mu^i$\,$\in$\,$\{0.1,0.3,0.5\}$, whereas non-random initializations are defined by centroids from a previous training run on class 0 (one epoch).
The latter is done to have a non-random centroid initialization that is as dissimilar as possible from the training data.
The initialization of the precisions cannot be varied, because empirical data shows that training converges to undesirable solutions if the precisions are not initialized to large values.
While this will have to be investigated further, we find that convergence to near-identical levels, regardless of centroid initialization for all datasets (see \cref{tab:init} for more details).
\subsection{Added Value of Annealing}\label{sec:exp:added}
To demonstrate the beneficial effects of annealing, we perform experiments on all datasets with annealing turned off.
This is achieved by setting $\sigma_0=\sigma_\infty$.
This invariably produces sparse-component solutions with strongly inferior log-likelihoods after training, please refer to \cref{tab:init}.
\newcolumntype{C}[1]{>{\centering\let\newline\\\arraybackslash\hspace{0pt}}b{#1}}
\newcolumntype{L}[1]{>{\raggedright\let\newline\\\arraybackslash\hspace{0pt}}b{#1}}
\setlength\tabcolsep{1pt}
\begin{table*}[htb!]
	\label{tab:init}
	\caption{
		Effect of different random and non-random centroid initializations of SGD training.
		Given are the means and standard deviations of final log-likelihoods (10 repetitions per experiment).
		To show the added value of annealing, the right-most column indicates the final log-likelihoods when annealing is turned off.
		This value should be compared to the leftmost entry in each row where annealing is turned on.
		Standard deviations in this case where very small so they are omitted.
	}
	\centering
	\scriptsize
	\begin{tabular}{|lL{.55cm}|l|l|l|l|l|l!{\vrule width 1pt}l|l!{\vrule width 1pt}c|}
		\hline
		\multicolumn{2}{|c|}{\multirow{3}{*}{\diagbox[width=55pt,height=25pt]{Dataset}{Initialization}}} &  \multicolumn{6}{c!{\vrule width 1pt}}{\textbf{random}}                                                                                                                                                                                                                             & \multicolumn{2}{c!{\vrule width 1pt}}{\textbf{non-random}}  & \tiny \textbf{no annealing}                                       \\
		&  &  \multicolumn{2}{c|}{$\mu^i$\,$=$\,$0.1$}                                           & \multicolumn{2}{c|}{$\mu^i$\,$=$\,$0.3$}                                             & \multicolumn{2}{c!{\vrule width 1pt}}{$\mu^i$\,$=$\,$0.5$}                                             & \multicolumn{2}{c!{\vrule width 1pt}}{init class 0}          & $\mu^i=0.1$                             \\
		&  &  \multicolumn{1}{c|}{\scriptsize$mean$}   & \multicolumn{1}{c|}{\scriptsize $std$}  & \multicolumn{1}{c|}{\scriptsize$mean$}    & \multicolumn{1}{c|}{\scriptsize $std$}   & \multicolumn{1}{c|}{\scriptsize $mean$}    & \multicolumn{1}{c!{\vrule width 1pt}}{\scriptsize $std$}  & \multicolumn{1}{c|}{\scriptsize$mean$}  & \multicolumn{1}{c!{\vrule width 1pt}}{\scriptsize $std$} & \scriptsize{$mean$}\\ \hline
		\multicolumn{2}{|l|}{MNIST}                                                                     &  \tablenum[table-format=5.2]{  205.47}    & \tablenum[table-format=3.2]{ 1.08}      & \tablenum[table-format=5.2]{  205.46}     & \tablenum[table-format=3.2]{  0.77}      & \tablenum[table-format=5.2]{  205.68}      & \tablenum[table-format=3.2]{  0.78}                       & \tablenum[table-format=5.2]{  205.37}   & \tablenum[table-format=4.2]{   0.68}  & 124.1 \\ \hhline{|-----------|}
		\multicolumn{2}{|l|}{FashionMNIST}                                                              &  \tablenum[table-format=5.2]{  231.22}    & \tablenum[table-format=3.2]{ 1.53}      & \tablenum[table-format=5.2]{  231.58}     & \tablenum[table-format=3.2]{  2.84}      & \tablenum[table-format=5.2]{  231.00}      & \tablenum[table-format=3.2]{  1.11}                       & \tablenum[table-format=5.2]{  229.59}   & \tablenum[table-format=4.2]{   0.59}   & 183.0\\ \hhline{|-----------|}
		\multicolumn{2}{|l|}{NotMNIST}                                                                  &  \tablenum[table-format=5.2]{  -48.41}    & \tablenum[table-format=3.2]{ 1.77}      & \tablenum[table-format=5.2]{  -48.59}     & \tablenum[table-format=3.2]{  1.56}      & \tablenum[table-format=5.2]{  -48.32}      & \tablenum[table-format=3.2]{  1.13}                       & \tablenum[table-format=5.2]{  -49.37}   & \tablenum[table-format=4.2]{   2.32}   & -203.8\\ \hhline{|-----------|}
		\multicolumn{2}{|l|}{Devanagari}                                                                &  \tablenum[table-format=5.2]{  -15.95}    & \tablenum[table-format=3.2]{ 1.59}      & \tablenum[table-format=5.2]{  -15.76}     & \tablenum[table-format=3.2]{  1.34}      & \tablenum[table-format=5.2]{  -17.01}      & \tablenum[table-format=3.2]{  1.11}                       & \tablenum[table-format=5.2]{  -22.07}   & \tablenum[table-format=4.2]{   4.59}   & -263.4\\ \hhline{|-----------|}
		\multicolumn{2}{|l|}{Fruits 360}                                                                &  \tablenum[table-format=5.2]{12095.80}    & \tablenum[table-format=3.2]{98.02}      & \tablenum[table-format=5.2]{12000.70}     & \tablenum[table-format=3.2]{127.00}      & \tablenum[table-format=5.2]{12036.25}      & \tablenum[table-format=3.2]{122.06}                       & \tablenum[table-format=5.2]{10912.79}   & \tablenum[table-format=4.2]{1727.61}   & 331.2\\ \hhline{|-----------|}
		\multicolumn{2}{|l|}{SVHN}                                                                      &  \tablenum[table-format=5.2]{ 1328.06}    & \tablenum[table-format=3.2]{ 0.94}      & \tablenum[table-format=5.2]{ 1327.99}     & \tablenum[table-format=3.2]{  1.59}      & \tablenum[table-format=5.2]{ 1328.40}      & \tablenum[table-format=3.2]{  1.17}                       & \tablenum[table-format=5.2]{ 1327.80}   & \tablenum[table-format=4.2]{   0.94}   & 863.2\\ \hhline{|-----------|}
		\multicolumn{2}{|l|}{ISOLET}                                                                    &  \tablenum[table-format=5.2]{  354.34}    & \tablenum[table-format=3.2]{ 0.04}      & \tablenum[table-format=5.2]{  354.36}     & \tablenum[table-format=3.2]{  0.04}      & \tablenum[table-format=5.2]{  354.36}      & \tablenum[table-format=3.2]{  0.04}                       & \tablenum[table-format=5.2]{  354.20}   & \tablenum[table-format=4.2]{   0.05}   & 201.5\\ \hline
	\end{tabular}
\end{table*}
\subsection{Clustering Performance Evaluation} 
To compare the clustering performance of sEM and GMM the Davies-Bouldin score~\cite{Davies1979} and the Dunn index~\cite{Dunn1973} are determined.
We evaluate the grid-search results to find the best parameter setup for each metric for comparison.
sEM is initialized by k-means to show that our approach does not depend on parameter initialization.
\Cref{tab:clustering_metrics} indicaties that SGD can egalize sEM performance.
\newcolumntype{C}[1]{>{\centering\let\newline\\\arraybackslash\hspace{0pt}}b{#1}}
\newcolumntype{L}[1]{>{\raggedright\let\newline\\\arraybackslash\hspace{0pt}}b{#1}}
\setlength\tabcolsep{4pt}
\renewcommand{\arraystretch}{0.99}
\begin{table}[htb!]
	\label{tab:clustering_metrics}
	\caption{
		Clustering performance comparison of SGD and sEM training using Davies-Bouldin score (less is better) and Dunn index (more is better).
		Each time mean metric value (of $10$ experiment repetitions) at the end of training, and their standard deviations are presented.
		Results are in bold face whenever they are better by more than half a standard deviation.
	}
	\centering
	\small	
	\begin{tabular}{|lL{.55cm}|l|l!{\vrule width 1pt}l|l!{\vrule width 1pt}l|l!{\vrule width 1pt}l|l|} 
		\hline
		\multicolumn{2}{|c|}{\multirow{3}{*}{\diagbox[width=65pt,height=27.5pt]{Dataset}{\shortstack[r]{Metric\\{\small Algo.}}}}}        &  \multicolumn{4}{c!{\vrule width 1pt}}{\textbf{Davies-Bouldin score}}                                                                                                                                             &  \multicolumn{4}{c|}{\textbf{Dunn index}}                                                                                                                                             \\
		&  &  \multicolumn{2}{c!{\vrule width 1pt}}{\textbf{SGD}}                                               & \multicolumn{2}{c!{\vrule width 1pt}}{\textbf{sEM}}                                                          &  \multicolumn{2}{c!{\vrule width 1pt}}{\textbf{SGD}}                                               & \multicolumn{2}{c|}{\textbf{sEM}}                                                \\
		&  &  \multicolumn{1}{c|}{\scriptsize$mean$}  & \multicolumn{1}{c!{\vrule width 1pt}}{\scriptsize$std$} & \multicolumn{1}{c|}{\scriptsize$mean$}           & \multicolumn{1}{c!{\vrule width 1pt}}{\scriptsize $std$}  &  \multicolumn{1}{c|}{\scriptsize$mean$}  & \multicolumn{1}{c!{\vrule width 1pt}}{\scriptsize$std$} & \multicolumn{1}{c|}{\scriptsize$mean$}  & \multicolumn{1}{c|}{\scriptsize $std$} \\ \hline
		\multicolumn{2}{|l|}{MNIST}                                                                                                     &  \tablenum[table-format=1.2]{2.50}       & \tablenum[table-format=1.2]{0.04}                       & \hfill             $\mathbf{2.47}$               & \tablenum[table-format=1.2]{0.04}                         &  \tablenum[table-format=1.2]{0.18}       & \tablenum[table-format=1.2]{0.02}                       & \tablenum[table-format=1.2]{0.16}       & \tablenum[table-format=1.2]{0.02}      \\ \hhline{|----------|}
		\multicolumn{2}{|l|}{FashionMNIST}                                                                                              &  \hfill             $\mathbf{2.06}$      & \tablenum[table-format=1.2]{0.05}                       & \tablenum[table-format=1.2]{2.20}                & \tablenum[table-format=1.2]{0.04}                         &  \tablenum[table-format=1.2]{0.20}       & \tablenum[table-format=1.2]{0.03}                       & \tablenum[table-format=1.2]{0.19}       & \tablenum[table-format=1.2]{0.02}      \\ \hhline{|----------|}
		\multicolumn{2}{|l|}{NotMNIST}                                                                                                  &  \tablenum[table-format=1.2]{2.30}       & \tablenum[table-format=1.2]{0.03}                       & \hfill             $\mathbf{2.12}$               & \tablenum[table-format=1.2]{0.03}                         &  \tablenum[table-format=1.2]{0.15}       & \tablenum[table-format=1.2]{0.03}                       & \tablenum[table-format=1.2]{0.14}       & \tablenum[table-format=1.2]{0.04}      \\ \hhline{|----------|}
		\multicolumn{2}{|l|}{Devanagari}                                                                                                &  \hfill             $\mathbf{2.60}$      & \tablenum[table-format=1.2]{0.04}                       & \tablenum[table-format=1.2]{2.64}                & \tablenum[table-format=1.2]{0.02}                         &  \hfill             $\mathbf{0.33}$      & \tablenum[table-format=1.2]{0.01}                       & \tablenum[table-format=1.2]{0.27}       & \tablenum[table-format=1.2]{0.04}      \\ \hhline{|----------|}
		\multicolumn{2}{|l|}{SVHN}                                                                                                      &  \hfill             $\mathbf{2.34}$      & \tablenum[table-format=1.2]{0.04}                       & \tablenum[table-format=1.2]{2.41}                & \tablenum[table-format=1.2]{0.03}                         &  \tablenum[table-format=1.2]{0.15}       & \tablenum[table-format=1.2]{0.02}                       & \tablenum[table-format=1.2]{0.15}       & \tablenum[table-format=1.2]{0.02}      \\ \hhline{|----------|}
	\end{tabular} 
	
\end{table}
\subsection{Streaming Experiments with Constant Statistics}\label{sec:exp:str}
\begin{figure}[htb!]
	\centering
	\subfigure[][MNIST]{
		\includegraphics[width=0.3\linewidth]{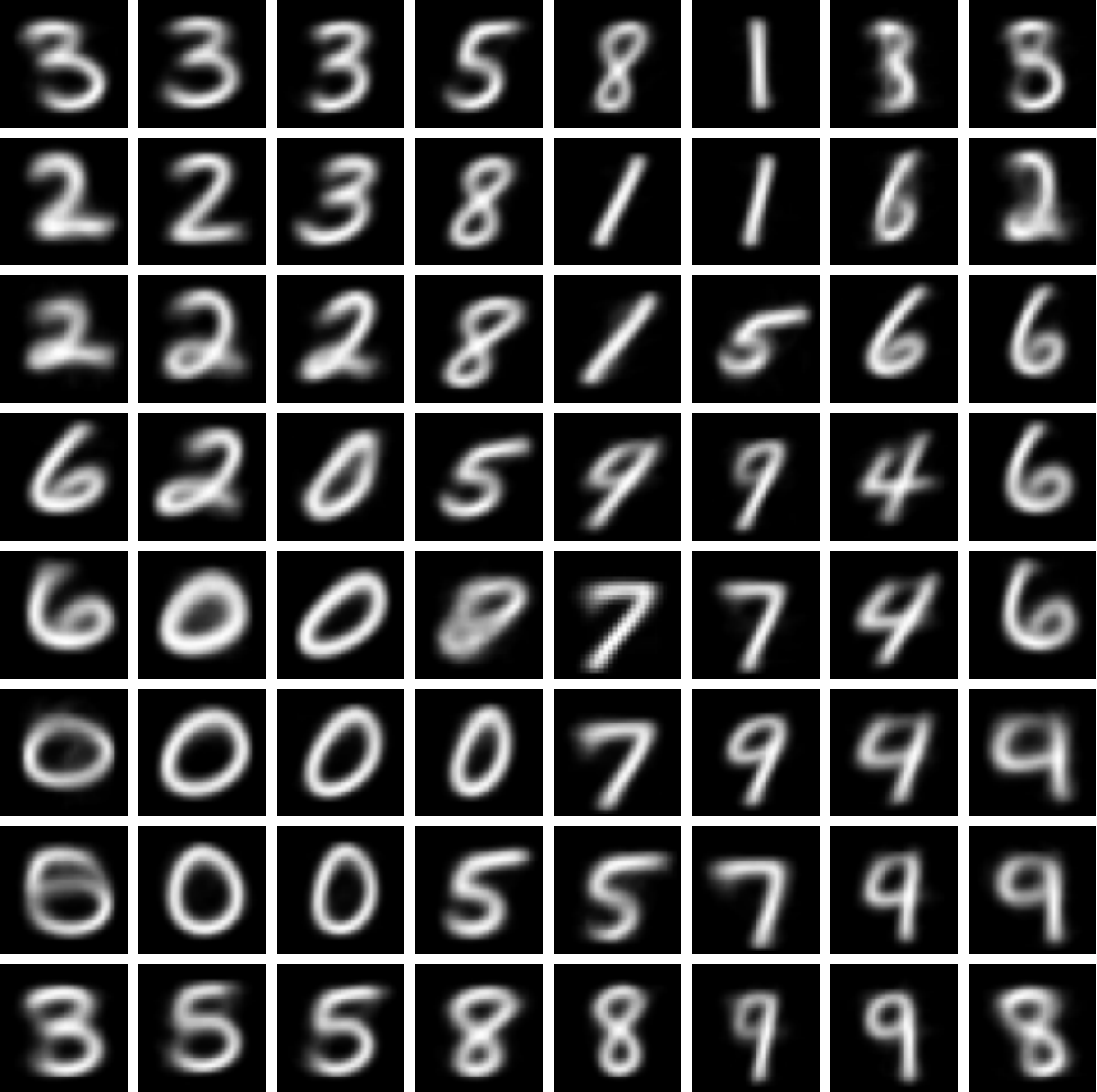}
		\label{fig:MNIST}
	}\subfigure[][SVHN]{
		\includegraphics[width=0.3\linewidth]{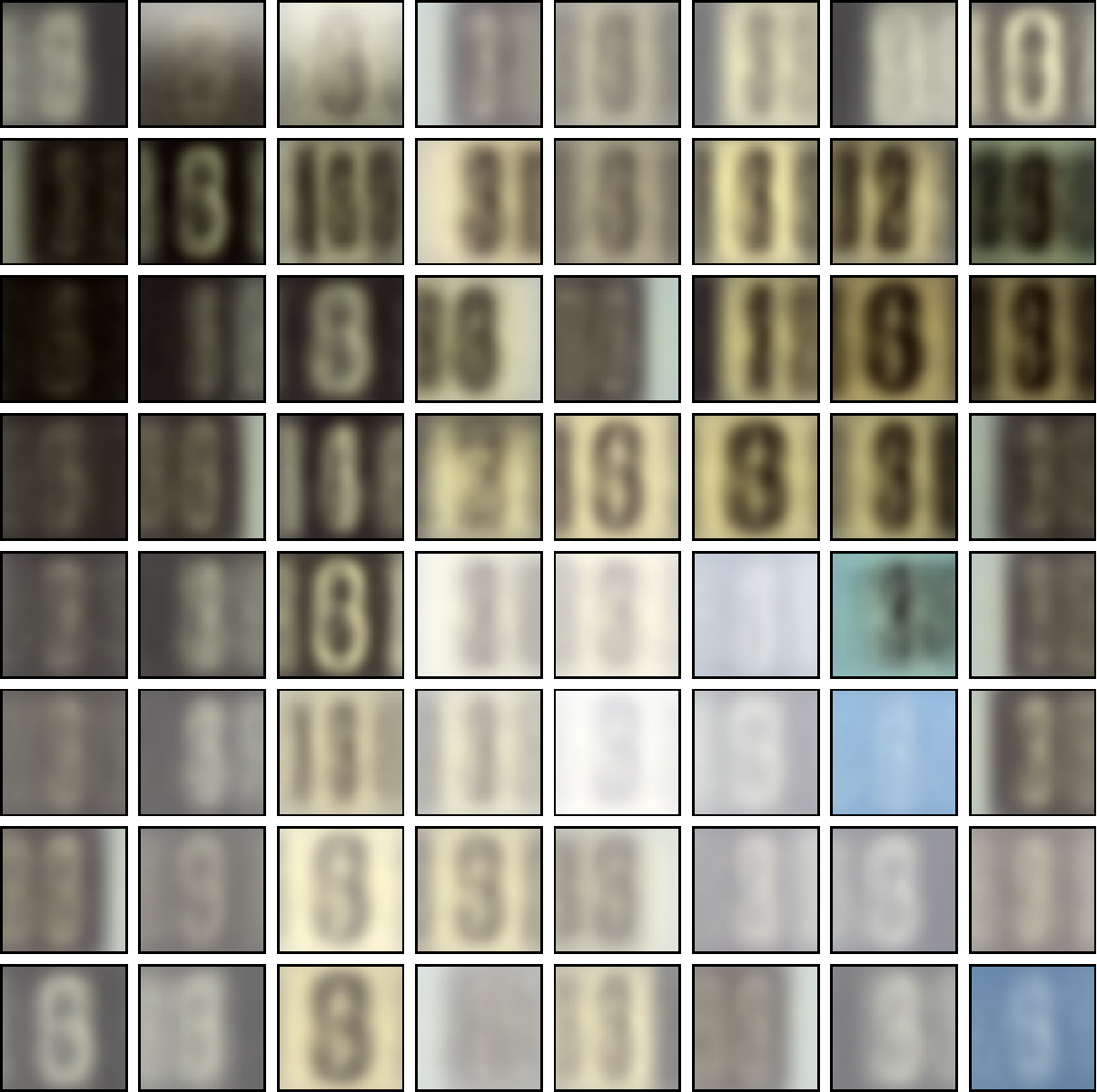}
		\label{fig:SVHN}
	}\subfigure[][FashionMNIST]{
		\includegraphics[width=0.3\linewidth]{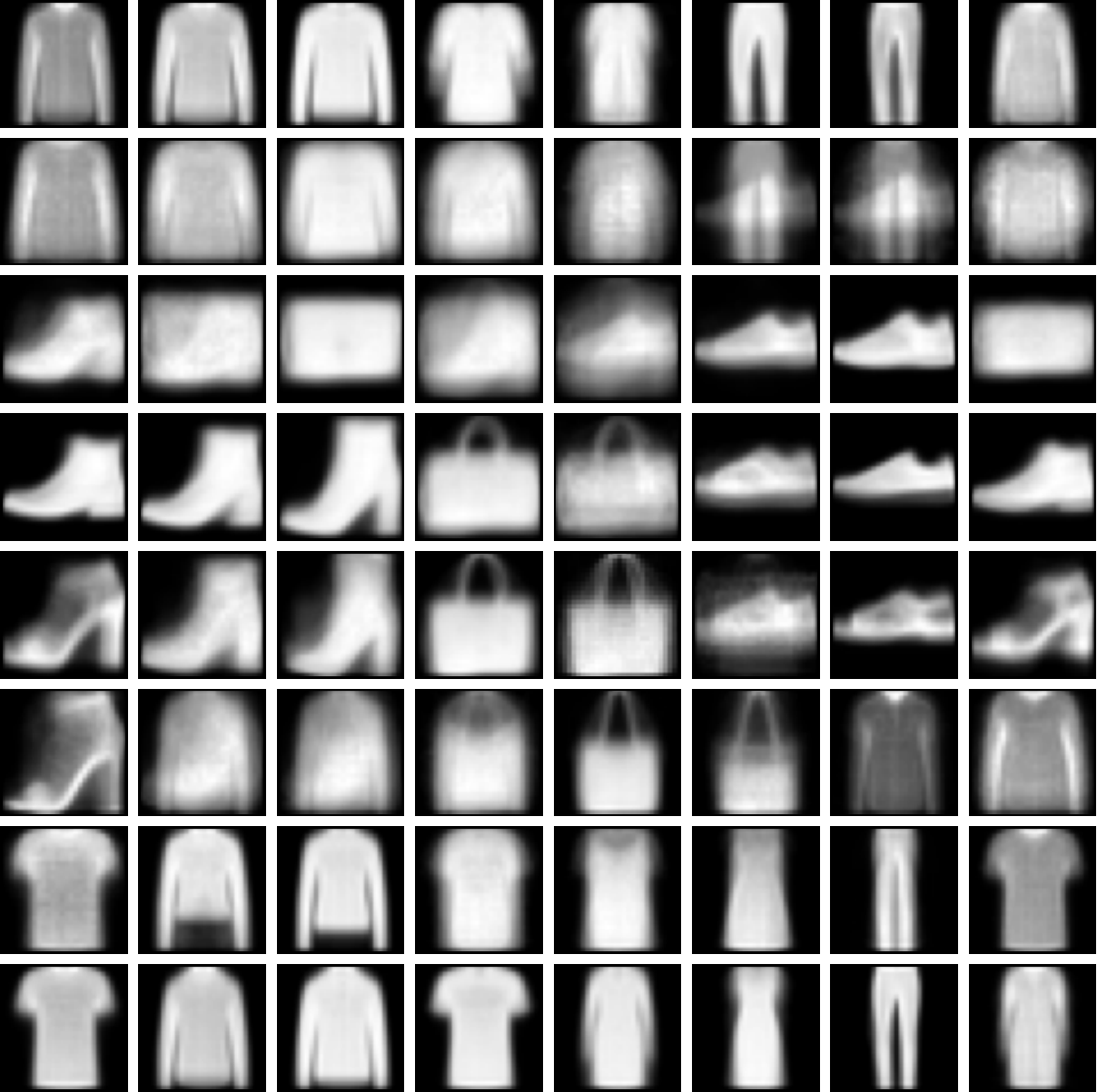}
		\label{fig:FashionMNIST}
	}
	\\
	\subfigure[][Devanagari]{
		\includegraphics[width=0.3\linewidth]{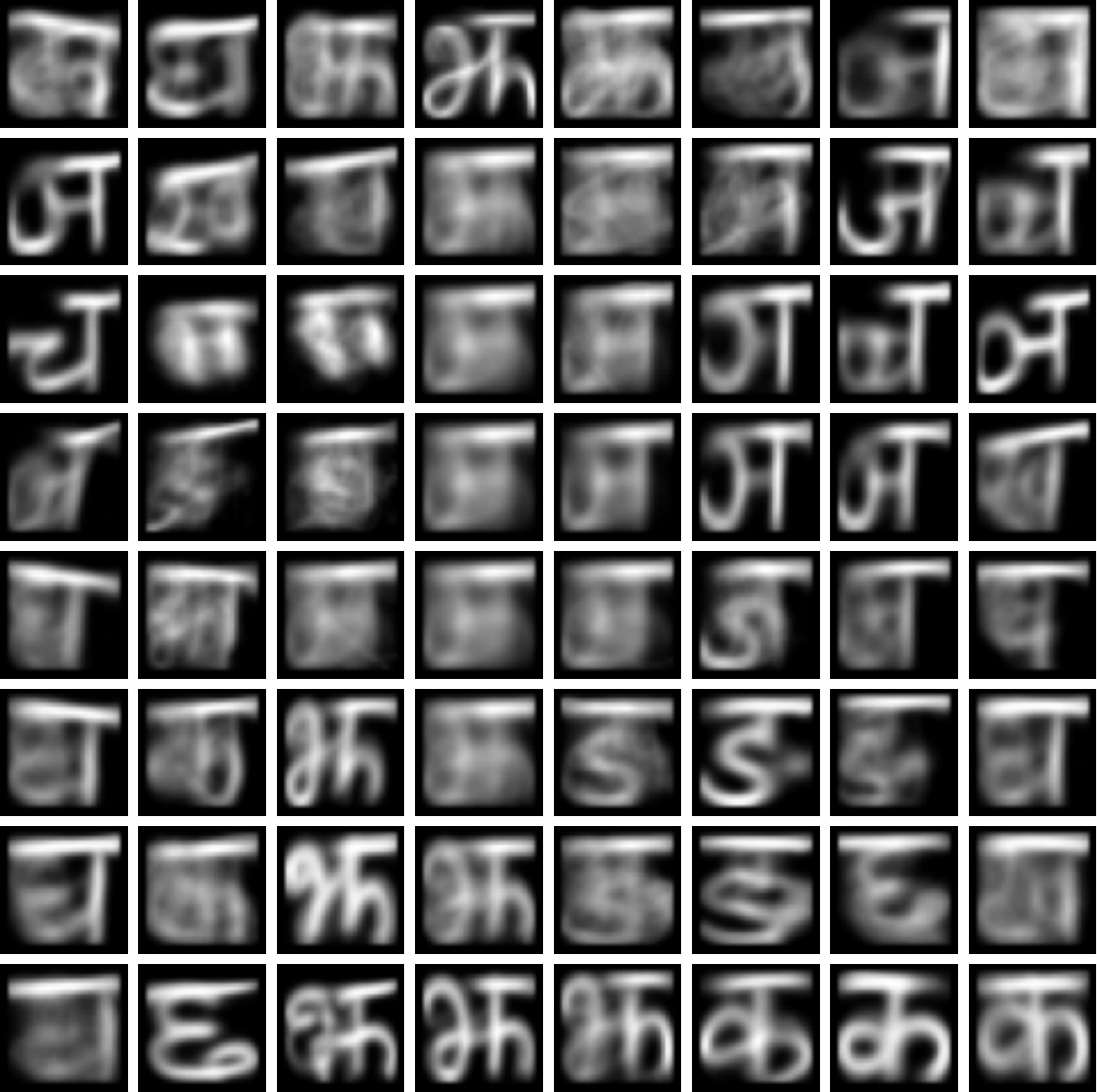}
		\label{fig:Devanagari}
	}\subfigure[][NotMNIST]{
		\includegraphics[width=0.3\linewidth]{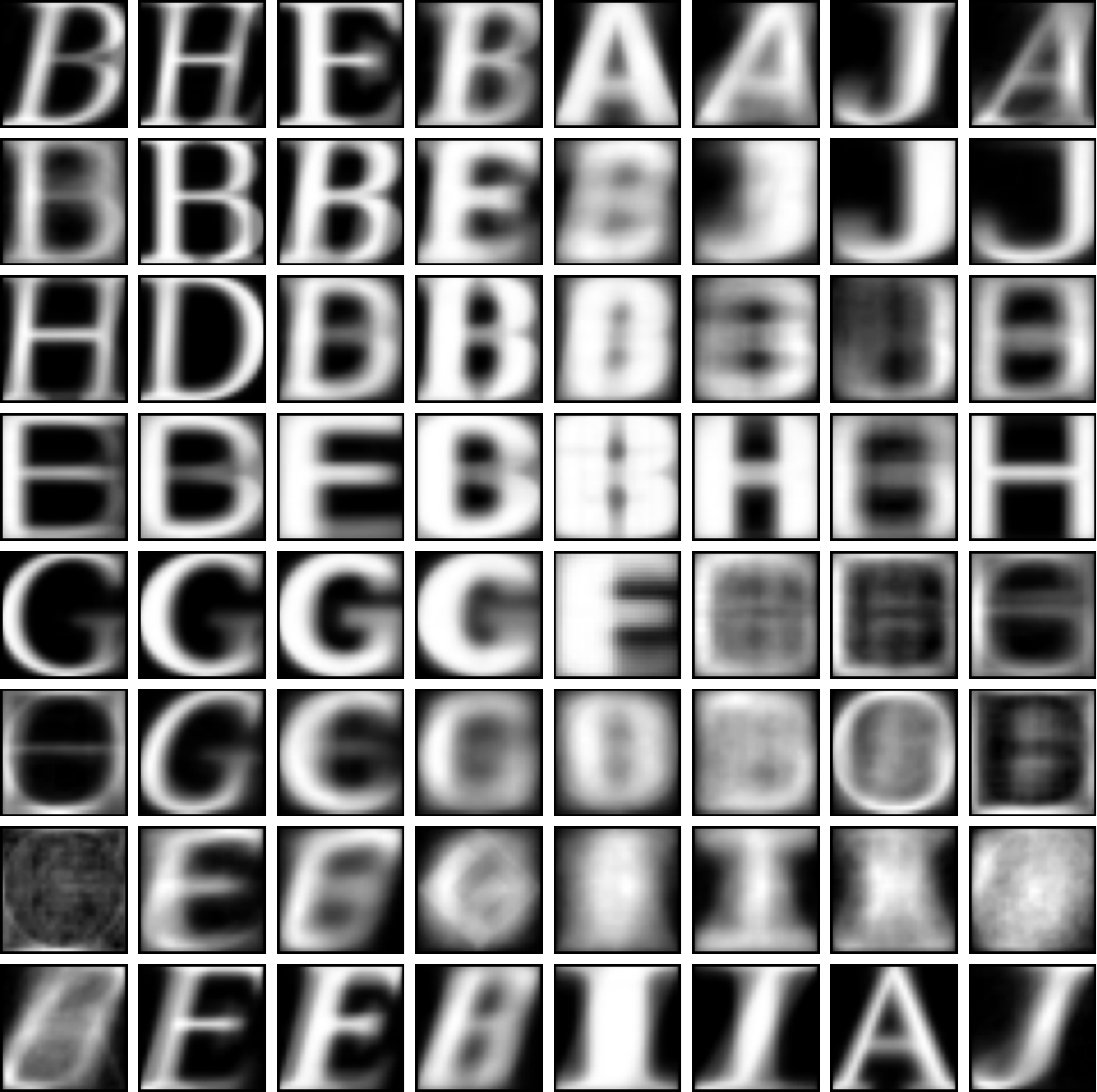}
		\label{fig:NotMNIST}
	}\subfigure[][Fruits 360]{
		\includegraphics[width=0.3\linewidth]{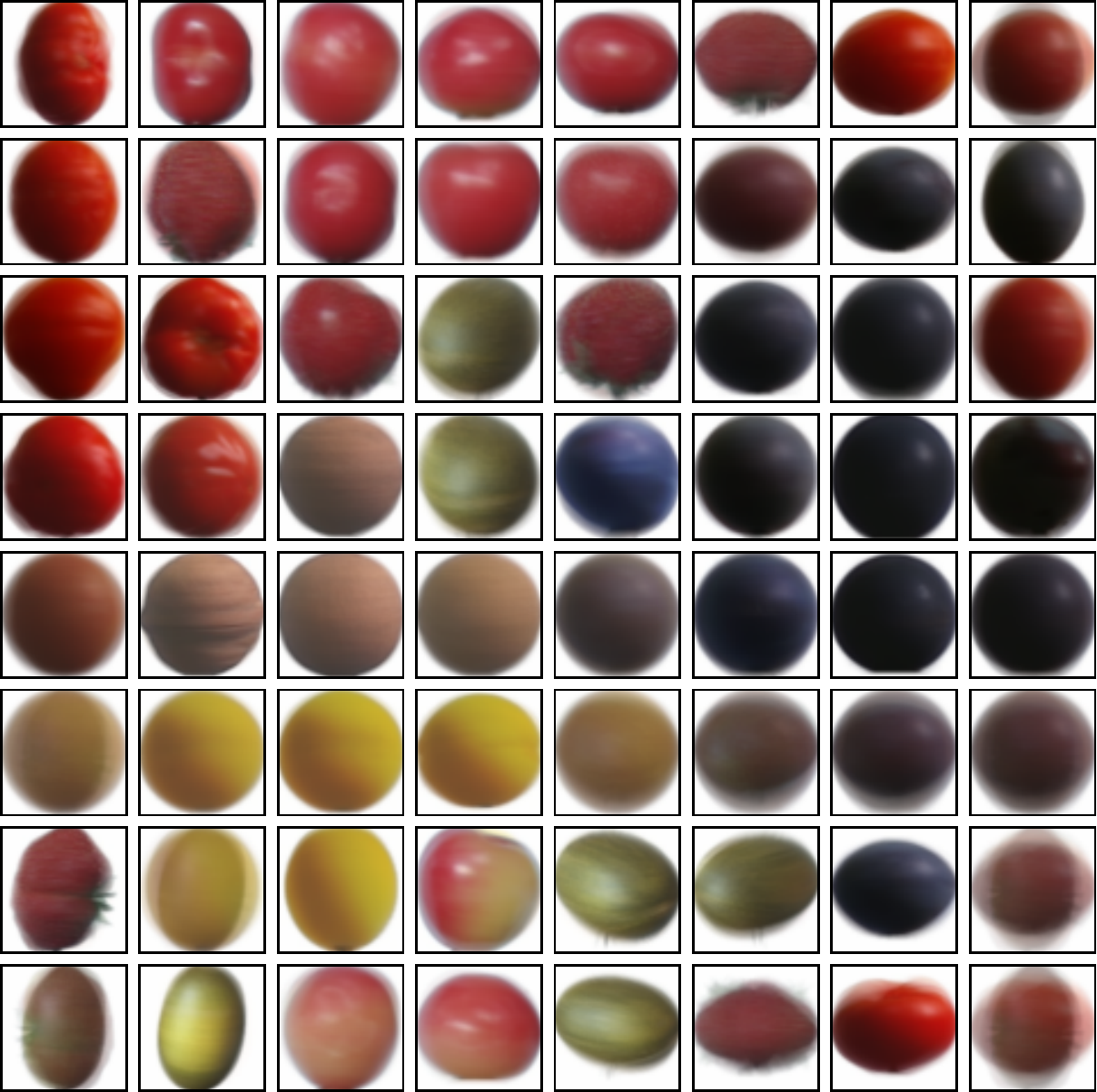}
		\label{fig:Fruits}
	}
	\caption{Exemplary results for centroids learned by SGD.\label{fig:examples1}}
\end{figure}

We train GMMs for three epochs (enough for convergence in all cases) using SGD and sEM on all datasets as described in \cref{sec:train-sem,sec:train-sgd}. 
The resulting centroids of our SGD-based approach are shown in \cref{fig:examples1}, whereas the final loss values for SGD and sEM are compared in \cref{tab:str}.
The centroids for both approaches are visually similar, except for the topological organization due to annealing for SGD, and the fact that in most experiments, some components do not converge for sEM while the others do.
\Cref{tab:str} indicates that SGD achieves performances superior to sEM in the majority of cases, in particular for the highest-dimensional datasets (SVHN: \num{3072} and Fruits 360: \num{30000} dimensions).
\newcolumntype{C}[1]{>{\centering\let\newline\\\arraybackslash\hspace{0pt}}b{#1}}
\newcolumntype{L}[1]{>{\raggedright\let\newline\\\arraybackslash\hspace{0pt}}b{#1}}
\setlength\tabcolsep{1pt}
\renewcommand{\arraystretch}{0.91}
\begin{table}[htb!]
	\label{tab:str}
	\caption{
		Comparison of SGD and sEM training on all datasets in a streaming-data scenario.
		Shown are log-likelihoods at the end of training, averaged over 10 repetitions, along with their standard deviations. 
		Results are in bold face whenever they are higher by more than half a standard deviation.
		Additionally, the averaged maximum responsibilities ($p_{k^*}$) for test data are given for justifying the max-component approximation.
	}
	\centering
	\small
	\begin{tabular}{|lL{.1cm}|l|l|l!{\vrule width 1pt}l|l|} 
		\hline
		\multicolumn{2}{|c|}{\multirow{2}{*}{\diagbox[width=55pt,height=17pt]{Dataset}{Algorithm}}}  &  \multicolumn{3}{c!{\vrule width 1pt}}{\textbf{SGD}}                                                                                                & \multicolumn{2}{c|}{\textbf{sEM}}                                                \\
		&  & \multicolumn{1}{c|}{$\varnothing\max p_{k^*}$}  & \multicolumn{1}{c|}{\scriptsize$mean$}    & \multicolumn{1}{c!{\vrule width 1pt}}{\scriptsize$std$} & \multicolumn{1}{c|}{\scriptsize$mean$}  & \multicolumn{1}{c|}{\scriptsize $std$} \\ \hline
		\multicolumn{2}{|l|}{MNIST}                                                                   & \tablenum[table-format=1.6]{0.992674}           & \tablenum[table-format=5.1]{  216.6}     & \tablenum[table-format=3.2]{ 0.31}                      & \tablenum[table-format=5.1]{ 216.8}    & \tablenum[table-format=4.2]{   1.38}   \\ \hhline{|-------|}
		\multicolumn{2}{|l|}{{\scriptsize FashionMNIST}}                                                            & \tablenum[table-format=1.6]{0.997609}           & \hfill             $\mathbf{  234.5}$    & \tablenum[table-format=3.2]{ 2.28}                      & \tablenum[table-format=5.1]{ 222.9}    & \tablenum[table-format=4.2]{   6.03}   \\ \hhline{|-------|}
		\multicolumn{2}{|l|}{NotMNIST}                                                                & \tablenum[table-format=1.6]{0.998713}           & \hfill             $\mathbf{  -34.7}$    & \tablenum[table-format=3.2]{ 1.16}                      & \tablenum[table-format=5.1]{ -40.0}    & \tablenum[table-format=4.2]{   8.90}   \\ \hhline{|-------|}
		\multicolumn{2}{|l|}{Devanagari}                                                              & \tablenum[table-format=1.6]{0.999253}           & \hfill             $\mathbf{  -14.6}$    & \tablenum[table-format=3.2]{ 1.09}                      & \tablenum[table-format=5.1]{ -13.4}    & \tablenum[table-format=4.2]{   6.16}   \\ \hhline{|-------|}
		\multicolumn{2}{|l|}{Fruits 360}                                                              & \tablenum[table-format=1.6]{0.999746}           & \hfill             $\mathbf{11754.3}$    & \tablenum[table-format=3.2]{75.63}                      & \tablenum[table-format=5.1]{5483.0}    & \tablenum[table-format=4.2]{1201.60}   \\ \hhline{|-------|}
		\multicolumn{2}{|l|}{SVHN}                                                                    & \tablenum[table-format=1.6]{0.998148}           & \hfill             $\mathbf{ 1329.8}$    & \tablenum[table-format=3.2]{ 0.80}                      & \tablenum[table-format=5.1]{1176.0}    & \tablenum[table-format=4.2]{  16.91}   \\ \hhline{|-------|}
		\multicolumn{2}{|l|}{ISOLET}                                                                  & \tablenum[table-format=1.6]{0.994069}           & \tablenum[table-format=5.1]{  354.2}     & \tablenum[table-format=3.2]{ 0.01}                      & \tablenum[table-format=5.1]{ 354.5}    & \tablenum[table-format=4.2]{   0.37}   \\ \hline
	\end{tabular}
\end{table}

\begin{figure}[htb!]
	\centering
	\label{fig:sem}
	\hfill\includegraphics[width=0.31\linewidth]{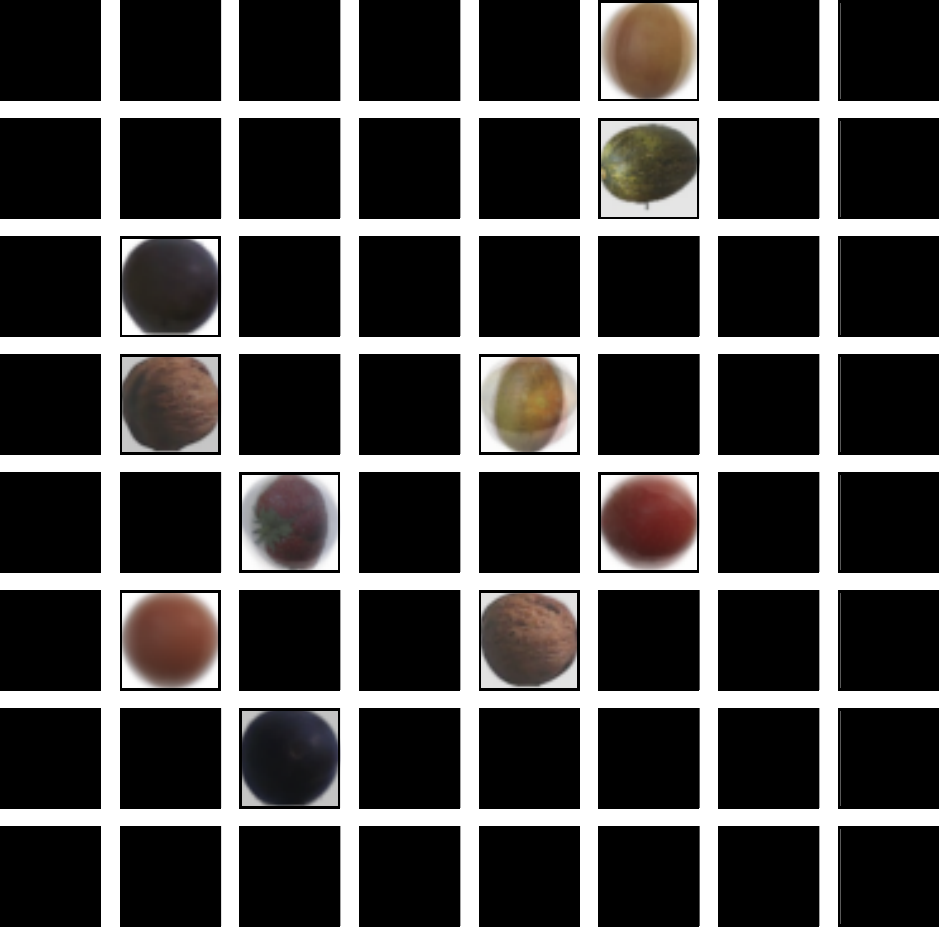}\hfill\includegraphics[width=0.31\linewidth]{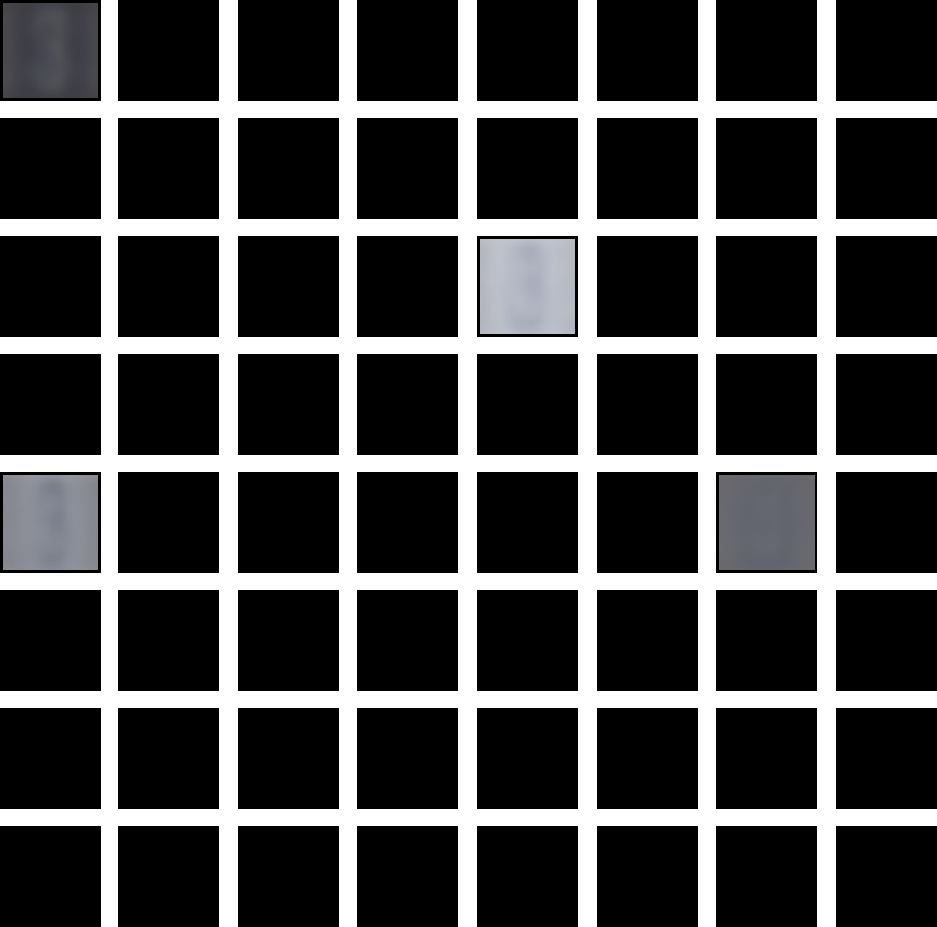}\hfill~
	\caption{
		Visualization of centroids after training runs ($3$ epochs) on high-dimensional datasets for sEM: Fruits 360 (left, \num{30000} dimensions) and SVHN (right, \num{3000} dimensions).
		Component entries are displayed \enquote{as is}, meaning that low brightness means low RGB values. 
		Many GMM components remain unconverged, which is analogous to a sparse-component solution and explains the low log-likelihood values for these high-dimensional datasets.
	}
\end{figure}

\paragraph{Visualization of High-dimensional sEM Outcomes}\label{app:sem}
\Cref{fig:sem} was obtained after training GMMs by sEM on both the Fruits 360 and the SVHN dataset.
It should be compared to \cref{fig:examples1}, where an identical procedure was used to visualize centroids of SGD-trained GMMs.
It is notable that the effect of unconverged components does not occur at all for our SGD approach, which is due to the annealing mechanism that \enquote{drags} unconverged components along.
\section{Assumptions made by EM and SGD}\label{sec:em_assumption}
The EM algorithm assumes that the observed data samples $\{\vx_n\}$ depend on unobserved latent variables $\vz_n$ in a non-trivial fashion.
This assumption is formalized for a GMM with K components by formulating the complete-data likelihood in which $\vz_n$\,$\equiv$\,$z_n\in\{0,\dots, K-1\}$ is a scalar:
\begin{equation}\label{eqn:complete}
	p(\vx_n,z_n) = \pi_{z_n}\mathcal{N}_{z_n}(\vx_n)
\end{equation}
where we have defined $\mathcal{N}_{k}(\vx_n)$\,$=$\,$\mathcal{N}(\vx_n;\vtheta_k,\vmu_k)$ for brevity.
It is assumed that the $z_n$ are unobservable random variables whose distribution is unknown.
Marginalizing them out gives us the incomplete-data likelihood $p(\vx_n)\!=\!\sum_k p(\vx_n,z_n)$.
The derivation of the EM algorithm starts out with the total incomplete-data log-likelihood
\begin{equation}
	\begin{split}
		\mathcal{L} & = \log p(X) = \log \prod_n p(\vx_n) = \sum_n \log p(\vx_n)    \\
		& = \sum_n \log \sum_k p(\vx_n,z_n=k)                           \\
		& = \sum_n \log \sum_k p(z_n=k)\frac{p(\vx_n,z_n=k)}{p(z_n=k)}.
	\end{split}
\end{equation}
Due to the assumption that $\mathcal{L}$ is obtained by marginalizing out the latent variables, an explicit dependency on $z_n$ can be re-introduced.
For the last expression, Jensen' inequality can be used to construct a lower bound:
\begin{equation}
	\begin{split}
		\mathcal{L} & \sim \sum_n \log \sum_{k} p(\vz_n=k) \frac{p(\vx_n,\vz_n=k)}{p(\vz_n=k)}               \\
		            & \ge \sum_n \sum_{k} p(\vz_n=k) \log \frac{p(\vx_n,\vz_n=k)}{p(\vz_n=k)}\label{eqn:lb}.
	\end{split}
\end{equation}
Since the realizations of the latent variables are unknown, we can assume any form for their distribution.
In particular, for the choice $p(z_n)\sim p(\vx_n,z_n)$, the lower bound becomes tight.
Simple algebra and the fact that the distribution $p(z_n)$ must be normalized gives us:
\begin{equation}
	\begin{split}
		p(z_n=k) & = \frac{p(z_n=k,\vx_n)}{p(\vx_n)}                                    \\
		         & = p(z_n=k|\vx_n)                                                     \\
		         & = \frac{p(z_n=k,\vx_n)}{\sum_l p(z_n=l,\vx_n)}                       \\
		         & = \frac{\pi_k\mathcal{N}_k(\vx_n)}{\sum_l \pi_l\mathcal{N}_l(\vx_n)}
	\end{split}
\end{equation}
where we have used \cref{eqn:complete} in the last step.
$p(z_n\!=\!k|\vx_n)$ is a quantity that can be computed from data with no reference to the latent variables.
For GMM it is usually termed \textit{responsibility} and we write it as $p(z_n\!=\!k|\vx_n)\!\equiv\!\gamma_{nk}$.
\par
However, the construction of a tight lower bound, which is actually different from $\mathcal{L}$, only works when $p(\vx_n,z_n)$ depends non-trivially on the latent variable $z_n$.
If this is not the case,
we have $p(\vx_n,z_n)$\,$=$\,$K^{-1}p(\vx_n)$ and the derivation of \cref{eqn:lb} goes down very differently:
\begin{equation}
	\begin{split}
		\mathcal{L} & \sim \sum_n \log p(\vx_n) \ge  \sum_n \sum_{k} p(z_n=k) \log \frac{p(\vx_n,z_n=k)}{p(z_n=k)} \\
		            & = \sum_n \sum_{k} p(z_n=k) \log \frac{K^{-1}p(\vx_n)}{p(z_n=k)}                              \\
		            & = \sum_n \log\Big(K^{-1}p(\vx_n)\Big) - \sum_{k} p(z_n=k) \log p(z_n=k)                      \\
		            & \equiv \sum_n \Big( \log p(\vx_n) - \big(\log K - \mathcal{H}[z_n]\big)\Big)
	\end{split}
	\raisetag{20pt}
\end{equation}
where $\mathcal{H}$ represents the Shannon entropy of $p(\vz)$.
The highest value this can have is $\log K$ for an uniform distribution of the $z_n$, finally leading to a lower bound for $\mathcal{L}$ of
\begin{equation}
	\mathcal{L} \ge  \sum_n \Big( \log p(\vx_n)\Big) 
\end{equation}
which is trivial by Jensen's inequality, but not tight.
In particular, no closed-form solutions to the associated extrema value problem can be computed.
\par
This shows that optimizing GMM by EM assumes that each sample has been drawn from a single element in a set of $K$ uni-modal Gaussian distributions.
Which distribution is selected for sampling depends on a latent random variable.
On the other hand, optimization by SGD uses the incomplete-data log-likelihood $\mathcal{L}$ as basis for optimization, without assuming the existence of hidden variables at all.
This may be advantageous for problems where the assumption of Gaussianity is badly violated, although empirical studies indicate that optimization by EM works very well in a very wide range of scenarios.

\section{Discussion and Conclusion}\label{sec:discussion} 
The \textbf{relevance of this article} is outlined by the fact that training GMMs by SGD was recently investigated in the community by \cite{Hosseini2015,hosseini2019}.
We go beyond, since our approach does not rely on off-line data-driven model initialization, and works for high-dimensional streaming data. 
The presented SGD scheme is simple and very robust to initial conditions due to the proposed annealing procedure, see \cref{sec:exp:init} and \cref{sec:exp:added}.
In addition, our SGD approach compares favorably to the reference model for online EM \cite{Cappe2009} in terms of achieved log-likelihoods, which was verified on multiple real-world datasets.
Superior SGD performance is observed for the high-dimensional datasets.
\smallskip\\
\textbf{Analysis of results} suggests that SGD performs better than sEM on average, see \cref{sec:exp:str}, although the differences are very modest.
It should be stated clearly that it cannot be expected, and is not the goal of this article, to outperform sEM by SGD in the general case, only to achieve a similar performance.
However, if sEM is used without, e.g., k-means initialization, components may not converge (see \cref{fig:sem}) for very high-dimensional data like Fruits 360 and SVHN datasets, which is why SGD outperforms sEM in this case.
Another important advantage of SGD over sEM is the fact that the only parameter that needs to be tuned is the learning rate $\epsilon$, whereas sEM has a complex and not intuitive dependency on $\rho_0$, $\rho_\infty$ and $\alpha_0$.
\smallskip\\
\textbf{Small batch sizes and streaming data} are possible with the SGD-based approach.
Throughout the experiments, we used a batch size of $1$, which allows streaming-data processing without the need to store any samples at all.
Larger batch sizes are possible and strongly increase execution speed.
In the conducted experiments, SGD (and sEM) usually converged within the first two epochs, which is a substantial advantage whenever huge sets of data have to be processed.
\smallskip\\
\textbf{No assumptions about data generation} are made by SGD in contrast to the EM and sEM algorithms.
The latter guarantees that the loss will not decrease due to an M-step.
This, however, assumes a non-trivial dependency of the data on an unobservable latent variable (shown in \cref{sec:em_assumption}).
In contrast, SGD makes no hard-to-verify assumptions, which is a rather philosophical point, but may be an advantage in certain situations where data are strongly non-Gaussian.
\smallskip\\
\textbf{Numerical stability} is assured by our SGD training approach.
It does not optimize the log-likelihood but its max-component approximation.
This approximation contains no exponentials at all, and is well justified by the results of \cref{tab:str} which shows that component probabilities are strongly peaked.
In fact, it is the gradient computations where numerical problems occurred, e.g., NaN values.
The \enquote{logsumexp} trick mitigates the problem, but does not eliminate it (see \cref{sec:max-comp}).
It cannot be used when gradients are computed automatically, which is what most machine learning frameworks do.
\smallskip\\
\textbf{Hyper-Parameter selection guidelines} are as follows:
the learning rate $\epsilon$ must be set by cross-validation (a good value is 0.001).
We empirically found that initializing precisions to the cut-off value $D_\text{max}$ and an uniform initialization of the $\pi_i$ are beneficial, and that centroids are best initialized to small random values.
A value of $D_\text{max}=20$ always worked in our experiments.
Generally, the cut-off must be much larger than the inverse of the data variance.
In many cases, it should be possible to estimate this roughly, even in streaming settings, especially when samples are normalized.
For density estimation, choosing higher values for $K$  leads to higher final log-likelihoods.
For clustering, $K$ should be selected using standard techniques for GMMs.
The parameter $\delta$ controls loss stationarity detection for the annealing procedure and was shown to perform well for $\delta$\,$=$\,$0.05$.
Larger values will lead to a faster decrease of $\sigma(t)$, which may impair convergence.
Smaller values are always admissible but lead to longer convergence times.
The annealing time constant $\alpha$ should be set to the GMM learning rate $\epsilon$ or lower.
Smaller values of $\alpha$ lead to longer convergence times since $\sigma(t)$ will be updated less often.
The initial value $\sigma_0$ needs to be large in order to enforce convergence for all components.
A typical value is $\sqrt{K}$.
The lower bound on $\sigma_\infty$ should be as small as possible in order to achieve high log-likelihoods (e.g., $0.01$, see \cref{ssec:proof_annealing_upper_bound} for a proof).
\section{Future Work}\label{sec:conclusion}
The presented work can be extended in several ways: 
First of all, annealing control could be simplified further by inferring good $\delta$ values from $\alpha$.
Likewise, \textit{increases} of $\sigma$ might be performed automatically when the loss rises sharply, indicating a task boundary.
As we found that GMM convergence times grow linear with the number of components, we will investigate hierarchical GMM models that operate like a Convolutional Neural Network (CNN), in which individual GMM only see a local patch of the input and can therefore have low $K$.
\section*{Conflict of Interest} 
The authors declare that they have no conflict of interest.

\bibliographystyle{spmpsci}      
\bibliography{mybib}   

The final authenticated version is published in \textit{Neural Processing Letters}.
\end{document}